\documentclass{llncs}

\usepackage{amsfonts}
\usepackage{graphicx}
\usepackage{multirow}

\usepackage{color}

\begin{document}
\title{BoxNet: Deep Learning Based Biomedical Image Segmentation Using Boxes Only Annotation}
\author{Lin Yang$^1$, Yizhe Zhang$^1$, Zhuo Zhao$^1$, Hao Zheng$^1$, Peixian Liang$^1$,\\ Michael T. C. Ying$^2$, Anil T. Ahuja$^3$, Danny Z. Chen$^1$}
\institute{$^1$ Department of Computer Science and Engineering,\\ University of Notre Dame, Notre Dame, IN 46556, USA\\
$^2$ Department of Health Technology and Informatics,\\ The Hong Kong Polytechnic University, Hung Hom, Hong Kong\\
$^3$ Department of Imaging and Interventional Radiology,\\
The Chinese University of Hong Kong, Prince of Wales Hospital,\\ Shatin, NT, Hong Kong}

\maketitle

\begin{abstract}
In recent years, deep learning (DL) methods have become powerful tools for biomedical image segmentation. However, high annotation efforts and costs are commonly needed to acquire sufficient biomedical training data for DL models.
To alleviate the burden of manual annotation, in this paper, we propose a new weakly supervised DL approach for biomedical image segmentation using boxes only annotation.
First, we develop a method to combine graph search (GS) and DL to generate fine object masks from box annotation, in which DL uses box annotation to compute a rough segmentation for GS and then GS is applied to locate the optimal object boundaries.
During the mask generation process, we carefully utilize information from box annotation to filter out potential errors, and
then use the generated masks to train an accurate DL segmentation network.
Extensive experiments on gland segmentation in histology images, lymph node segmentation in ultrasound images, and fungus segmentation in electron microscopy images show that our approach attains superior performance over the best known state-of-the-art weakly supervised DL method and is able to achieve (1) nearly the same accuracy compared to fully supervised DL methods with far less annotation effort, (2) significantly better results with similar annotation time, and (3) robust performance in various applications.
\end{abstract}

\section{Introduction}
\begin{figure}[t]
	\centering
	\includegraphics[width=11cm]{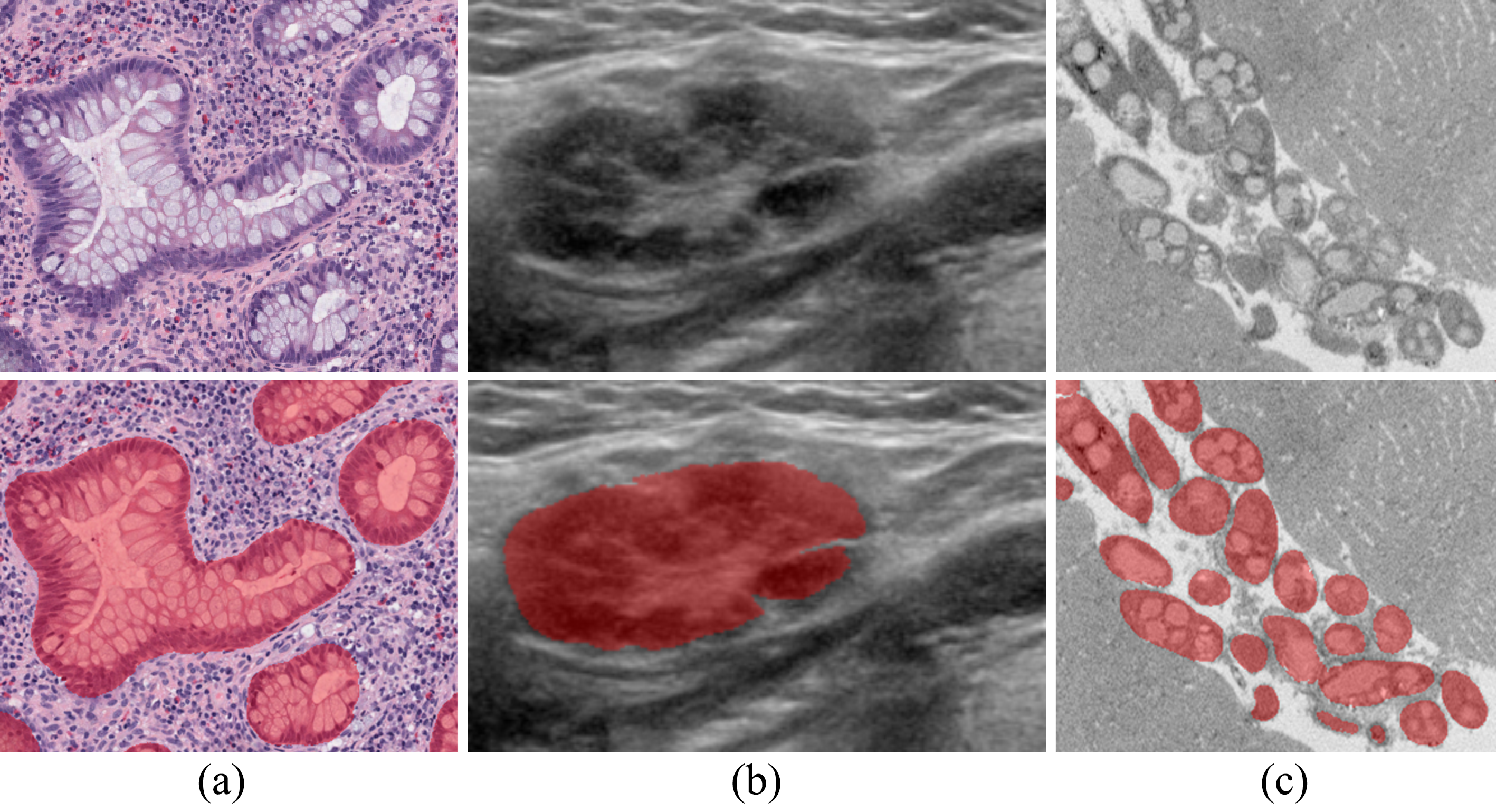}
	\caption[ ]{Example results computed by our method. The first row gives the original images and the second row shows our corresponding segmentation results (red masks) overlaid with the original images. These demonstrate that our method can achieve accurate segmentation and boundary delineation using boxes only annotation. (a) An H\&E stained histology image of glands; (b) an ultrasound image of lymph node; (c) an electron microscopy (EM) image of fungal cells.}
	\label{examples}
\end{figure}

In recent years, deep learning (DL) methods \cite{chen2016dcan,chen2016combining,ronneberger2015u} have become powerful tools for biomedical image segmentation.
However, due to large variety of biomedical applications (e.g., different targets, different imaging modalities, different experimental settings, etc), high annotation efforts and costs are commonly needed to acquire sufficient training data for DL models for new applications.
In biomedical image segmentation, studies have been done on reducing annotation effort by utilizing unannotated data~\cite{baur2017semi,zhang2017deep} and on annotation data selection~\cite{yang2017suggestive}.
In this paper, we present a different approach to alleviate the burden of manual annotation. Instead of using fine object masks to train a DL model, we propose a new weakly supervised DL approach that can achieve accurate segmentation by using only bounding boxes of the target objects as input.
See Fig.~\ref{examples} for some example results.
As annotating bounding boxes is $\sim10$ times faster than annotating fine masks \cite{papadopoulos2017extreme}, our approach can significantly reduce annotation effort.

A set of weakly supervised annotation methods has been proposed for semantic segmentation in natural scene images. In these methods, various weak annotation forms were explored (e.g., points \cite{bearman2016s}, scribbles \cite{lin2016scribblesup}, and bounding boxes \cite{dai2015boxsup,khoreva2017simple,papandreou2015weakly}).
Comparing to other weak annotation forms, bounding boxes are more well-defined to annotate and provide much more information (e.g., object sizes, more exhaustive background annotation).
Thus, among these weak annotation forms, bounding box approaches show the most promising results that could potentially match the segmentation results from full annotation \cite{khoreva2017simple}.

\begin{figure}[t]
	\centering
	\includegraphics[width=12cm]{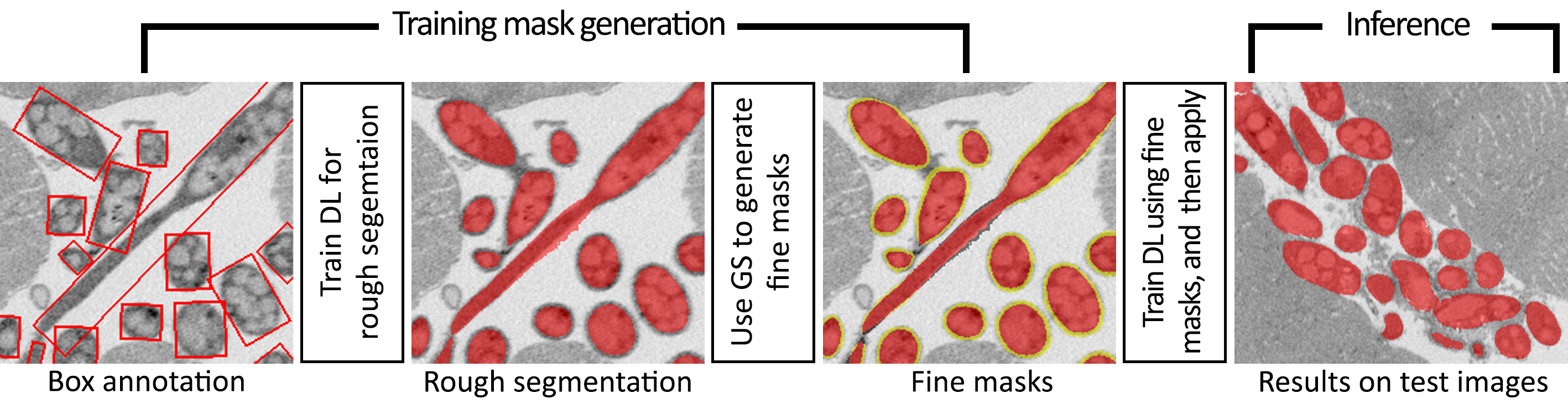}
	\caption[ ]{Illustrating the overall process of our approach. The example images are EM images of fungal cells.}
	\label{outline}
\end{figure}

However, the bounding box based methods for natural scene images cannot be directly extended to biomedical image segmentation for the following reasons.
(1) All these methods \cite{dai2015boxsup,khoreva2017simple,papandreou2015weakly} use orthogonal bounding boxes to annotate objects. It works well in natural scene settings in which many objects are often orthogonal to the image boundaries. But in biomedical images, objects can appear in any orientations and orthogonal bounding boxes are less useful (e.g., see Fig.~\ref{outline}). 
(2) Objects in biomedical images usually have more complicated inner structures and/or vague boundaries. Thus, the boundary recovery step (e.g., DenseCRF \cite{krahenbuhl2011efficient}) in these methods may not work well (Fig.~\ref{crfandgs}(c)).

Hence, in this paper, we develop a new bounding box based weakly supervised DL approach to deal with the two aforementioned challenges in biomedical image segmentation.
(1) To address the orthogonal bounding box issue, we present a method to efficiently annotate tilted bounding boxes based on the extreme points of target objects \cite{papadopoulos2017extreme}. Instead of a series of interactions (drawing bounding box, adjusting tilted angle, and adjusting box boundary), our method needs only six clicks for each bounding box (two clicks around the object center to indicate the box's orientation and four clicks for the extreme points on the object boundary), as shown in Fig.~\ref{annotation}. 
This greatly enhances the annotation efficiency and all these clicks can be reused in a later stage to better indicate object extents.
(2) To recover the object boundaries more accurately, instead of using the methods designed for natural scene images, we apply graph search (GS) \cite{li2006optimal}, a long-tested method for optimizing boundaries in biomedical images.
Fig.~\ref{outline} outlines the main ideas of our approach.
First, we develop a method to combine GS and DL to generate fine object masks from box annotation, in which DL uses box annotation to compute a rough segmentation for GS and GS is applied to locate the optimal object boundaries.
Note that, a key requirement of GS is to have a rough segmentation with correct topology. Our approach satisfies this requirement easily by using the topology information provided by the box annotation.
During the mask generation process, we carefully utilize information from box annotation to filter out potential errors, and
then use the generated masks to train an accurate DL network for image segmentation.

Experiments on gland segmentation in H\&E stained histology images \cite{zhang2016seeding}, lymph node segmentation in ultrasound images \cite{zhang2016coarse}, and fungus segmentation in electron microscopy (EM) images \cite{zhang2017deep} show that our approach attains superior performance over the best known state-of-the-art weakly supervised DL method \cite{khoreva2017simple}, and is able to achieve (1) nearly the same accuracy compared to fully supervised DL methods with far less annotation effort, (2) significantly better results with similar annotation time, and (3) robust performance in various applications.

\section{Method}

Fig.~\ref{outline} gives an overview of our approach. In this section, we focus on discussing three major components of our approach: (1) a procedure for annotating tilted bounding boxes; (2) a DL network to compute rough segmentation for graph search (GS) \cite{li2006optimal} based on box annotation; (3) fine mask generation using GS.

\subsection{A new procedure for annotating tilted bounding boxes \label{rbox}}
We first briefly review the known methods for annotating bounding boxes, and then present our new method for efficiently annotating tilted bounding boxes.

A standard protocol for annotating orthogonal bounding boxes in natural scene image datasets (e.g., ImageNet) usually has two steps \cite{papadopoulos2017extreme}: (1) Draw an orthogonal box by diagonally dragging the mouse from one corner of an imaginary rectangle that tightly bounds the object to the opposite corner; (2) adjust the boundary of the box until it actually bounds the object. Step (2) is often necessary, since the two corners annotated in step (1) may not be on the object, and it is quite challenging to annotate them accurately to align well with the object boundary. This standard protocol is difficult to extend to annotating tilted bounding boxes because it is much more time-consuming and difficult to draw a tilted rectangle than an orthogonal one as in step (1).
In \cite{papadopoulos2017extreme}, a new way for annotating orthogonal bounding boxes was proposed, which took only four clicks on the extreme points (top, bottom, leftmost, and rightmost) of the object to annotate the box. Since the extreme points are well-defined 
physical
points on the object boundary, it is much easier to accurately locate them than drawing an imaginary rectangle as in the standard protocol. The extreme point approach achieves $\sim 5$ times speedup comparing to the standard protocol \cite{papadopoulos2017extreme}.

\begin{figure}[t]
	\centering
	\includegraphics[width=12cm]{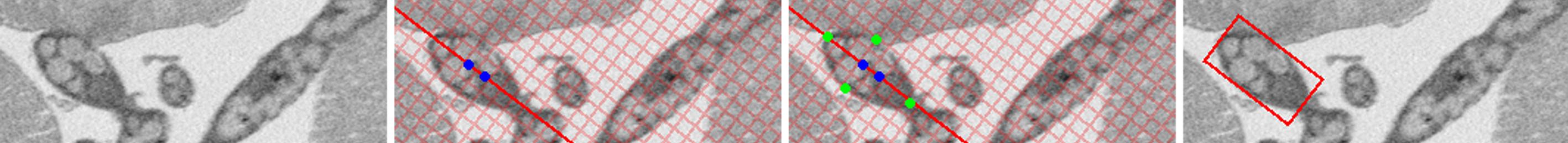}
	\caption[ ]{Illustrating our procedure for annotating tilted bounding boxes. The blue points represent the first two clicks that indicate the box's orientation; the green points represent the four extreme points with respect to the object's orientation. The example images are EM images of fungal cells.}
	\label{annotation}
\end{figure}

We adopt the extreme point approach and extend it to annotating tilted bounding boxes, for two advantages. (I) The extreme point approach not only is more efficient than the standard protocol, but also provides more information (e.g., where the object touches the bounding box). In Section~\ref{preseg}, we show how such extra information can help DL networks to generate a more accurate rough segmentation. (II) The extreme point approach can be easily extended to annotating tilted bounding boxes since the only required change is to click the extreme points with respect to the respective orientation of the object.

Fig.~\ref{annotation} shows the procedure for annotating a tilted bounding box in our approach. First, two clicks are used to annotate the orientation of the tilted box. To make every click count, these two clicks should be around the center of the object. We show how to utilize these two clicks in Section~\ref{preseg}. After the orientation of the box is acquired, we
draw an assistive grid (Fig.~\ref{annotation}) to help the user to annotate the four extreme points (top, bottom, leftmost, and rightmost) with respect to the object's orientation, in four clicks. Finally, the corresponding tilted bounding box of the extreme points is recorded (together with all the six clicks) and drawn on the original image to avoid duplicated annotation.

\subsection{Computing rough segmentation based on box annotation \label{preseg}}
To generate accurate fine object masks from box annotation, graph search (GS) needs the DL model to provide a rough segmentation that has the correct object topology and reasonable shape accuracy. In this section, we discuss how to make full use of all the information we acquire in Section~\ref{rbox}.

From the box annotation, we can gather the following cues.
(1) Since every object should be covered by at least one bounding box, the regions that are not covered by any boxes are expected to be the background.
(2) Each box is expected to contain one major object and the center of that object is specified by the first two clicks of the box annotation (Section~\ref{rbox}).
(3) The object is expected to touch the box on the four extreme points (Section~\ref{rbox}).

\begin{figure}[t]
	\centering
	\includegraphics[width=12cm]{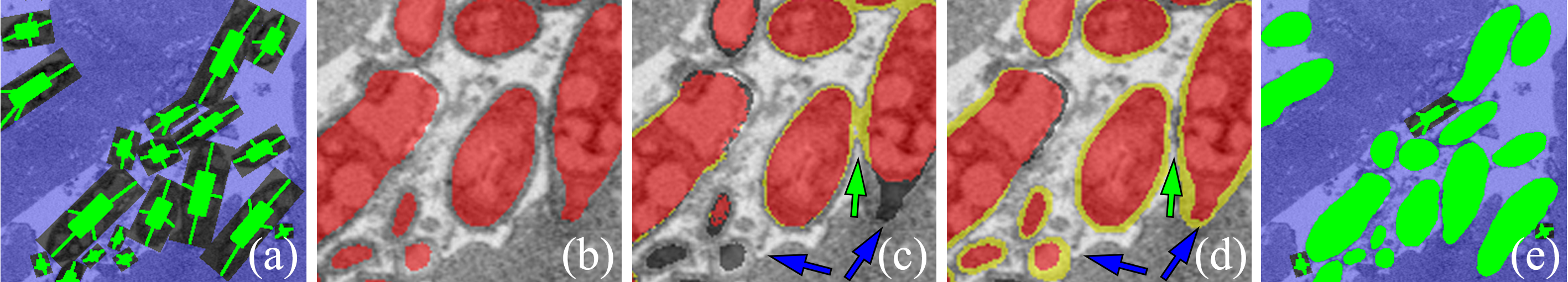}
	\caption[ ]{Illustrating the process of generating \textit{fine ground truth} from the \textit{box ground truth}. (a) An example of the \textit{box ground truth}; (b) rough segmentation results based on the \textit{box ground truth}; (c) and (d) show the refined results by DenseCRF \cite{krahenbuhl2011efficient} and graph search, respectively. Yellow (black) color represents the added (removed) areas after refinement. Blue (green) arrows point to the areas that are incorrectly removed (connected) by DenseCRF; (e) an example of generated \textit{fine ground truth}. The example images are EM images of fungal cells.}
	\label{crfandgs}
\end{figure}

Based on these cues, we generate the ground truth as shown in Fig.~\ref{crfandgs}(a). To distinguish this ground truth from the ground truth in Section~\ref{mask}, we call this ground truth the \textit{box ground truth}. The discussion in this paragraph refers to Fig.~\ref{crfandgs}(a). We take the following steps to label pixels in the images. (I) Based on cue (1), we label all the pixels not covered by any boxes as the background (blue color). 
(II) To promote the DL network to learn correct topology, based on cue (2), we label the pixels ``around'' the object's center as the object's class (green color).
Because the shape of a box is usually a good indicator for the shape of its object, we formally define the pixels ``around'' the object center as those pixels that are inside the rectangle which (a) is $k\%$ of the size of its bounding box, (b) has the same orientation as its bounding box, and (c) is centered at the object's center. 
The value of $k$ is related to the overall shapes of the target objects. For example, a more convex shape would allow a larger $k$. In all our experiments, we use $k=40\%$.
(III) To ensure reasonable shape accuracy, based on cue (3), four line segments are used to connect the extreme points and the center of the object (see Fig.~\ref{crfandgs}(a)).
Pixels on these four line segments are also labeled as the object's class (green color). This can better inform the DL model of the objects' extents, and it is especially important for the objects that have more than one layer of boundaries (e.g., glands), since this is the only information indicating which boundary layer should be detected.
(IV) The remaining unlabeled pixels (black color) 
are ignored during the training process by assigning a weight of 0.

Finally, a fully convolutional network (FCN) following the structure in \cite{yang2017suggestive} is used to compute a rough segmentation for GS based on the box ground truth. Fig.~\ref{crfandgs}(b) gives an example of the computed rough segmentation, and Table~\ref{pre-table} shows that a better rough segmentation can be obtained when all these cues are utilized. A possible issue for the above scheme is that it tends to work well with objects of relatively ``simple" shapes (e.g., star-shape).
However, our method is still broadly applicable in practice, since, on one hand, segmentation targets of a significant portion in biomedical images (e.g., the three applications in this paper) are star-shaped, and on the other hand, to handle more complex shapes, one can simply divide the object into multiple star-shaped regions and still annotate them using the current scheme (which would still be more efficient than tracing the objects).

\begin{figure}[t]
	\centering
	\includegraphics[width=\linewidth]{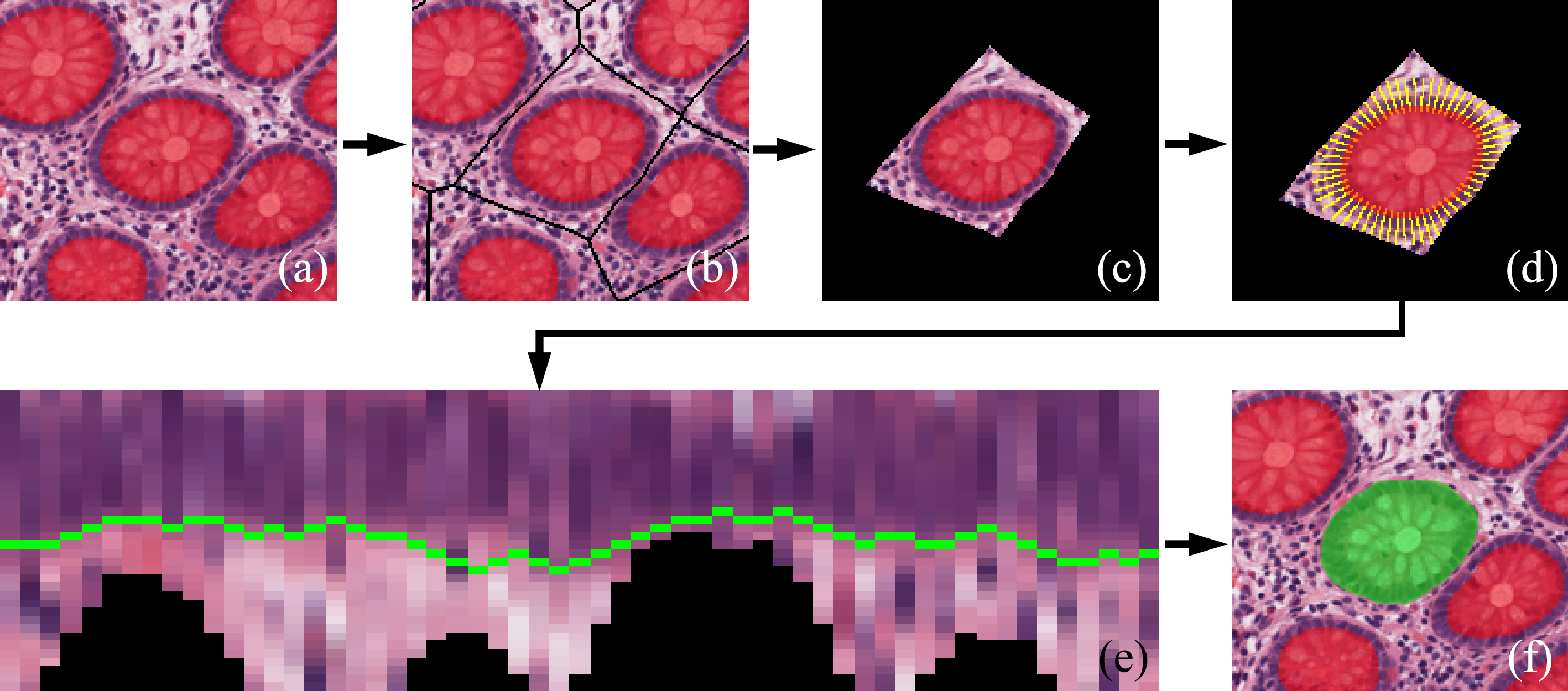}
	\caption[ ]{Illustrating the process of graph search (GS). (a) Rough segmentation (red masks); (b) the medial axis of the rough segmentation (black curves); (c) extracting the domain of a current object based on the medial axis; (d) image resampling and graph construction (yellow lines); (e) unfolded graph representation of the raw image (the green curve is the optimal object boundary computed by GS); (f) refined segmentation of (c) (green mask). The example images are H\&E stained histology images of glands.}
	\label{gsprocess}
\end{figure}

\subsection{Generating accurate masks from rough segmentation \label{mask}}

Although the rough segmentation computed in Section~\ref{preseg} is quite close to the results from fully supervised DL methods (see Tables~ \ref{final-table} and \ref{pre-table}), to bridge the final gap, we need to carefully utilize the local boundary information. A straightforward choice is to use DenseCRF \cite{krahenbuhl2011efficient} to promote better boundary delineation (as in \cite{khoreva2017simple}). However, as shown in Fig.~\ref{crfandgs}(c), DenseCRF does not work well in some biomedical images (especially the ones that have objects with complicated inner structures and/or vague boundaries). To better utilize the local boundary information, we show below how to address this issue using GS \cite{li2006optimal}.

\begin{table}[t]
	\centering
	\setlength{\tabcolsep}{2pt}
	\renewcommand{\arraystretch}{1.2}
	\caption{Comparison of final segmentation results. ``Full" represents \# of fully annotated images; ``weak" represents \# of images that are annotated only using boxes.}
	\begin{tabular}{| c || c | c | c || c | c | c || c | c | c |}
	  \hline
	  \multirow{2}{*}{Method} & \multicolumn{3}{c||}{Gland} & \multicolumn{3}{c||}{Lymph node} & \multicolumn{3}{c|}{Fungus} \\
	  \cline{2-10}
	  & Full & Weak & F1 & Full & Weak & F1 & Full & Weak & F1  \\
	  \hline
	  Similar annotation time & 2 & 0 & 0.9450 & 37 & 0 & 0.8740 & 4 & 0 & 0.9423 \\
	  \hline
	  Our method & 0 & 7 & 0.9617 & 0 & 170 & 0.9209 & 0 & 44 & 0.9607 \\
	  \hline
	  $Box^i$  \cite{khoreva2017simple} & 0 & 7 & 0.9124 & 0 & 170 & 0.8884 & 0 & 44 & 0.9267 \\
	  \hline
	  Fully supervised & 7 & 0 & 0.9648 & 170 & 0 & 0.9265 & 44 & 0 & 0.9608 \\
	  \hline
	\end{tabular}
	\label{final-table}
\end{table}

Comparing to DenseCRF, GS is more suitable for the task of generating accurate masks from rough segmentation, for the following reasons.
(1) GS does not change object topology (even though topology improvement may be desired in some applications). In our method, since we already obtain the topology from the box annotation, not changing the topology (by GS) is what we need for this problem (Fig.~\ref{crfandgs}(d)).
(2) 
Since GS ensures global optimal solutions, it can handle more complicated situations (e.g., when part of the boundary is vague or missing) in biomedical images.
(3) The parameters in GS have physical meanings which make them more intuitive to set across different applications.

Our method for producing accurate masks has two main steps.

\textbf{Step~1.} We pair each annotated box with a rough segmentation mask based on the Intersection over Union (IoU) score between the annotated box and the tilted bounding box of the rough segmentation mask (computed by using the same tilted angle as the annotated box).
Each annotated box is matched with a rough segmentation mask with the maximum IoU. To filter out potential errors, we only use GS to compute the fine masks for those annotated boxes that have matching rough segmentation masks with an $\mbox{IoU}\geq0.5$.

\textbf{Step~2.} The fine masks are then computed from the rough segmentation masks using GS. 
Fig.~\ref{gsprocess} illustrates the process of GS. First, to prevent overlapping masks, the medial axis of the rough segmentation is computed and used to ensure GS working on separated objects, as in \cite{liu2013optimal} (see (a)-(c) in Fig.~\ref{gsprocess}). Then, the graph construction process of GS follows the method in \cite{guo2018deep} in which the boundaries of the rough segmentation masks are used to determine the positions and directions of the graph columns (the reader is referred to \cite{guo2018deep} for more technical details). The cost function is simply the magnitude of gradients of the image intensities along the column directions (see (d)-(e) in Fig.~\ref{gsprocess}). Additionally, since the extreme points determined in Section~\ref{rbox} are on the boundaries of the objects and the bounding boxes should contain the segmentation, the boundaries generated by GS are forced to pass through the extreme points and forbidden to go outside the extents of the bounding boxes. These constraints are implemented by assigning very low (high) weights for pixels that should be excluded (included).

Finally, we generate a new ground truth based on the fine masks computed by GS. We call this ground truth the \textit{fine ground truth}. For all annotated boxes that have corresponding GS-generated masks, the \textit{box ground truth} is replaced by the generated masks. In all other locations, we keep using the \textit{box ground truth}. Fig.~\ref{crfandgs}(e) gives an example of the \textit{fine ground truth}. An FCN with the same structure as that in Section~\ref{preseg} is then trained using the \textit{fine ground truth} to produce accurate segmentation. See Fig.~\ref{examples} for some example results of segmentation.

\begin{table}[t]
	\centering
	\renewcommand{\arraystretch}{1.2}
	\caption{Comparison of rough segmentation results generated by different bounding box annotations. The results are evaluated using pixel-level F1 score.}
	\begin{tabular}{| c | c | c | c |}
	  \hline
	  Method & Gland & Lymph node & Fungus \\
	  \hline
	  Tilted bounding box + extreme points & \textbf{0.9513} & \textbf{0.914} & \textbf{0.9549} \\
	  \hline
	  Orthogonal bounding box \cite{khoreva2017simple} & 0.9101 & 0.8779 & 0.8992 \\
	  \hline
	  Orthogonal bounding box + extreme points & 0.9369 & 0.9087 & 0.929 \\
	  \hline
	\end{tabular}
	\label{pre-table}
\end{table}

\begin{table}[t]
	\centering
	\caption{Comparison of different GS usages. The results are evaluated using pixel-level F1 score.}
	\begin{tabular}{| c | c | c | c |}
	  \hline
	  Method & Gland & Lymph node & Fungus \\
	  \hline
	  Rough segmentation & 0.9513 & 0.914 & 0.9549 \\
	  \hline
	  Rough segmentation + GS & 0.9571 & 0.9153 & 0.9543 \\
	  \hline
	  Our method & \textbf{0.9617} & \textbf{0.9209} & \textbf{0.9607} \\
	  \hline
	\end{tabular}
	\label{GS-table}
\end{table}

\begin{figure}[t]
	\centering
	\includegraphics[width=11.45cm]{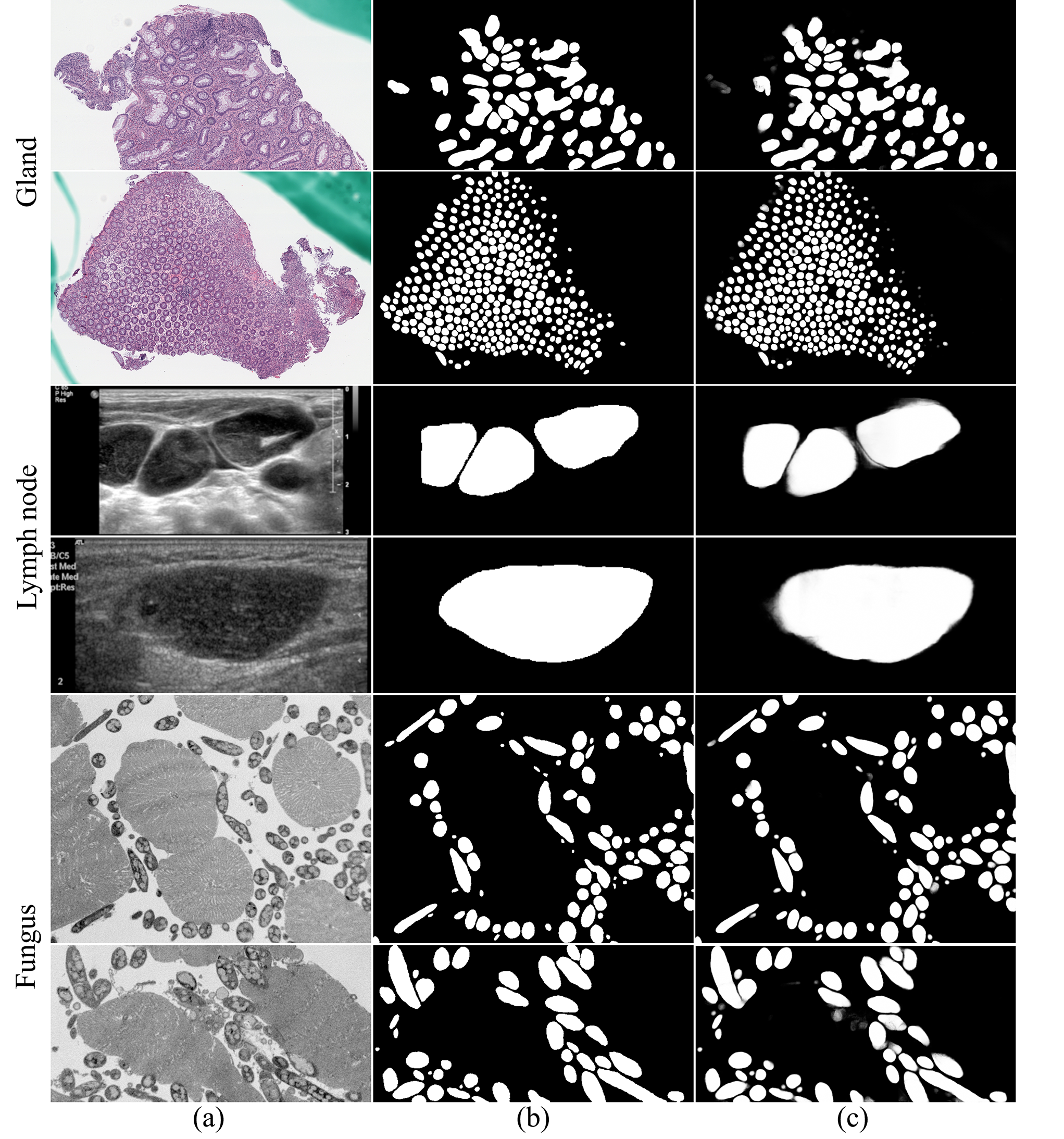}
	\caption[ ]{Some visual examples of our segmentation results. (a) The original images; (b) the ground truth; (c) the final segmentation results of our method.}
	\label{visual_examples}
\end{figure}

\section{Evaluation datasets and implementation details}
To thoroughly validate our method, in our experiments, we use three different datasets from various biomedical applications: (1) gland segmentation in H\&E stained histology images \cite{zhang2016seeding}, (2) lymph node segmentation in ultrasound images \cite{zhang2016coarse}, and (3) fungal cell segmentation in electron microscopy (EM) images \cite{zhang2017deep}.

\textbf{Gland segmentation dataset.} This dataset consists of 14 whole-slide clinical H\&E stained histology images of human intestinal tissues in various disease conditions (e.g., normal, chronic inflammation, acute inflammation, and chronic $+$ acute inflammation). We use 7 of them for training and the rest of them for testing. One might wonder whether 7 training images are too few to train our deep learning model. But we may keep in mind that whole-slide images usually have very large field of view (e.g., $14\mbox{k} \times 15\mbox{k}$ pixels) and each image can contain hundreds of glands. In our training set, there are in total 1058 glands, which is comparable to the 2015 MICCAI Gland Challenge dataset \cite{sirinukunwattana2017gland} (766 glands).

\textbf{Lymph node segmentation dataset.} This dataset contains 207 clinical ultrasound images of human neck lymph nodes. There are five types of lymph nodes (i.e., healthy, lymphoma, metastasis, reactive, and tuberculosis). We use 170 images for training and the remaining 37 for testing. 

\textbf{Fungus segmentation dataset.} This dataset contains 84 images captured by serial block-face scanning electron microscopy (EM). We use 44 images for training and the other 40 images for testing. 

\textbf{Implementation details.}
For all the three datasets, we rescale the intensity to $[0,1]$. Since the sizes of the training images can be much larger than the input size of the network, during each iteration, we form the training batch by randomly cropping the training images. After that, standard rotation and flipping of data augmentation are applied to the cropped patches. We use Adam optimizer with $\beta_1 = 0.9$, $\beta_2 = 0.999$, and $\epsilon = \mbox{1e-10}$ to train our network. The initial learning rate is set as $\mbox{5e-4}$ and reduced to $\mbox{5e-5}$ after 10k iterations. Our FCN components are trained for 20k iterations with a batch size of 8. Finally, the trained FCNs are applied to the test images in the same way as U-Net \cite{ronneberger2015u}.

\section{Experiments and results}

\begin{figure}[t]
	\centering
	\includegraphics[width=\linewidth]{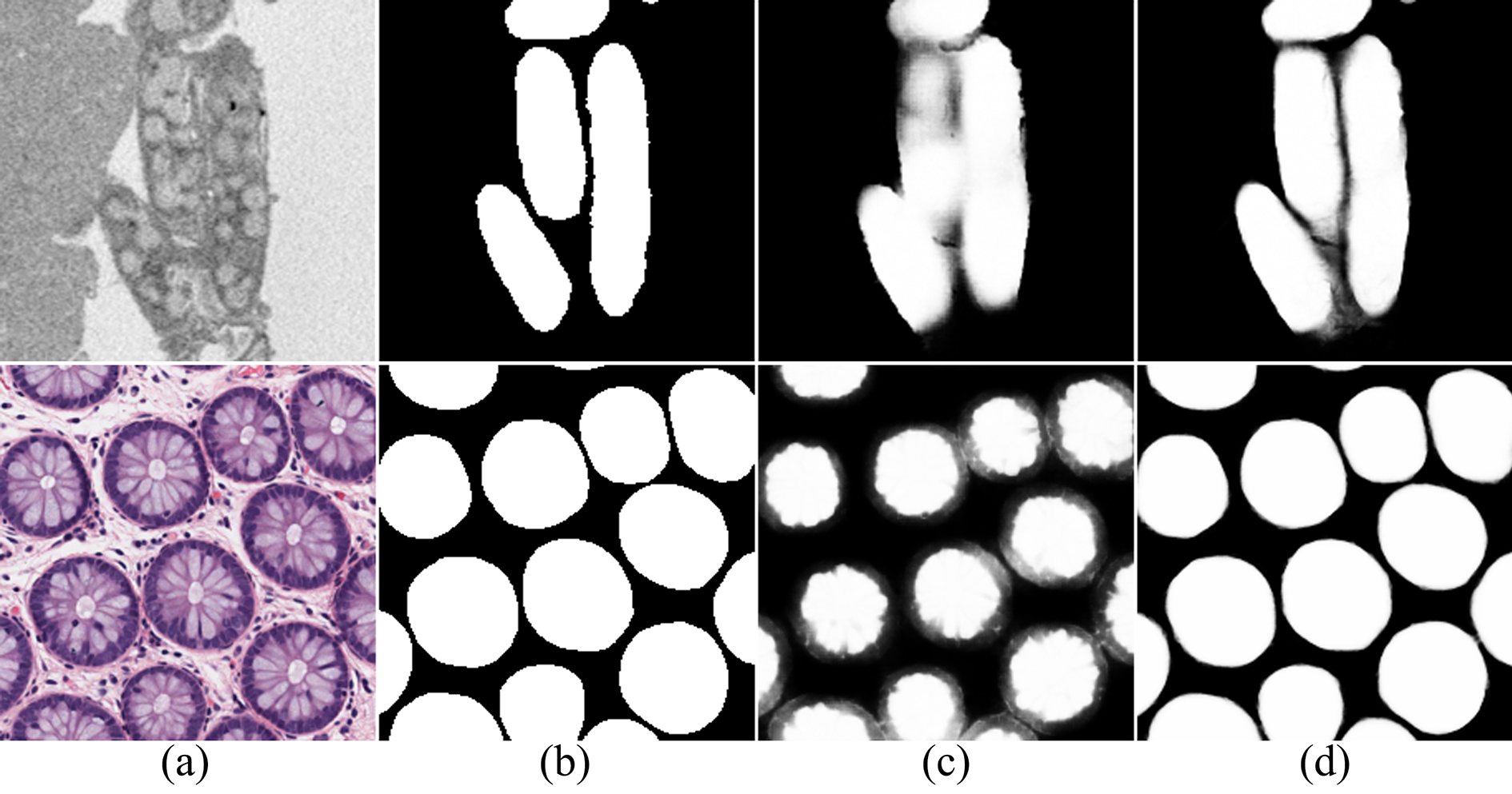}
	\caption[ ]{Qualitative comparison between our method and $Box^i$ \cite{khoreva2017simple}. (a) The original images; (b) the ground truth; (c) the final segmentation result of $Box^i$ \cite{khoreva2017simple}; (d) the final segmentation result of our method. The example images in the top row are EM images of fungal cells and the example images in the bottom row are H\&E stained histology images of glands.}
	\label{boxsf1}
\end{figure}

\subsection{Evaluation of our final segmentation results}
Fig.~\ref{visual_examples} shows some visual examples of our final segmentation results.
We evaluate the final segmentation of our method in three different aspects. (1) We compare our method with the best-known state-of-the-art weakly supervised DL method \cite{khoreva2017simple} using boxes only annotation. We choose the $Box^i$ variant in \cite{khoreva2017simple} to compare, since (a) the $GrabCut+$ and the $M\cap G+$ variant require supervised boundary detection that is not available to the three datasets we use, and (b) the $Box^i$ variant shows better results than \cite{dai2015boxsup,papandreou2015weakly}. (2) We compare our method with the same DL network trained on full annotation. (3) We compare our method with the same DL network trained on a subset of full annotation that takes similar annotation time as our box annotation. Table~\ref{final-table} shows that our approach attains superior performance over the best-known 
weakly supervised DL method \cite{khoreva2017simple}, and is able to achieve (I) nearly the same accuracy compared to fully supervised DL methods in far less annotation effort, and (II) much better results with similar annotation time.

We further provide some qualitative examples to demonstrate the effectiveness of our approach. In  the first row of Fig.~\ref{boxsf1}, one can see that, by preserving the topology of the box annotation using graph search (GS) \cite{li2006optimal}, our method can achieve much better object-level accuracy on the test data.
Furthermore, as shown in the second row of Fig.~\ref{boxsf1}, when the objects have more than one layer of boundaries, $Box^i$ \cite{khoreva2017simple} may fit any one of them while our method can detect the correct boundary layer by utilizing the cues from the extreme points.

\subsection{Ablation study}

\begin{figure}[t]
	\centering
	\includegraphics[width=12cm]{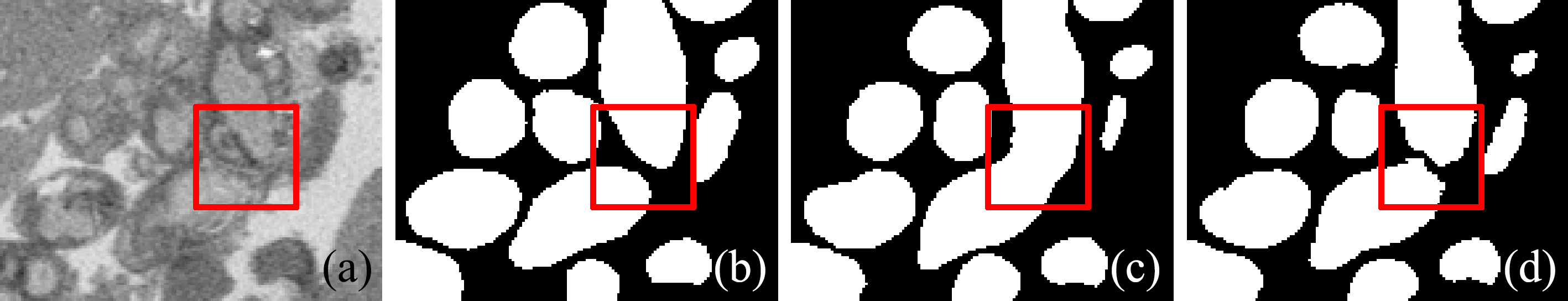}
	\caption[ ]{Qualitative comparison between our method and rough segmentation + GS. (a) The original image; (b) the ground truth; (c) rough segmentation + GS; (d) the final segmentation result of our method. The red boxes highlight a difficult image region for segmentation. The example images are EM images of fungal cells.}
	\label{gs-examples}
\end{figure}

\textbf{Different bounding box annotations.}
We evaluate the accuracy of the rough segmentation results produced by boxes only annotation. To show that tilted bounding boxes and the cues from extreme points are essential for our biomedical objects, we compare our approach with orthogonal bounding boxes only and orthogonal bounding boxes together with extreme points. The rough segmentation results of these methods are compared with the ground truth masks and evaluated using pixel-level F1 score. To evaluate their potential of being refined to be accurate masks, all the rough segmentation masks are dilated/eroded until they reach the maximum F1 score. As Table~\ref{pre-table} shows, by utilizing tilted bounding boxes and the cues from extreme points, our rough segmentation is much better than those of the other two methods and is only $\sim1\%$ worse than the full annotation (see Table~\ref{final-table}).

\textbf{Rough segmentation + GS \textit{vs.} our approach.}
As discussed in Section \ref{mask}, in our framework, we first use GS to refine the rough segmentation on the training images and then train a second FCN based on the refined results to generate the final segmentation. Yet, a more common and straightforward approach is to use GS as a post-processing step to refine the rough segmentation on the test images (we refer to this approach as ``rough segmentation + GS'' in Table~\ref{GS-table} and Fig.~\ref{gs-examples}).

Comparing to ``rough segmentation + GS'', our framework has the following advantages.
(1) Applying GS to the test images can only achieve cosmetic improvements. On the other hand, by providing more accurate boundary annotation (produced by GS) on the training images, our second FCN can detect object instances more accurately (see Fig.~\ref{gs-examples} and Table~\ref{GS-table}).
(2) GS could deteriorate the rough segmentation when the object topology is incorrect (see the fungus experiments in Table~\ref{GS-table}). By applying GS to the training images, we can filter out such potential errors using the box annotation. Furthermore, the extreme points can provide strong regulation on GS as well. Thus, our framework achieves consistently better results than ``rough segmentation + GS'' in all our datasets (Table~\ref{GS-table}).
(3) By shifting GS from the test images to the training images, our second FCN is able to mimic the behaviours of GS. Hence, there is no need for an additional GS-based post-processing step on test images in our framework, which can improve the inference speed.

\section{Conclusions}
In this paper, we presented a new weakly supervised DL approach for biomedical image segmentation using boxes only annotation that can achieve nearly the same performance compared to fully supervised DL methods. Our new method provides a more efficient way to annotate training data for biomedical image segmentation applications, and can potentially save considerable manual efforts.

 \subsubsection{Acknowledgment.}
This research was supported in part by NSF Grants CCF-1617735 and CNS-1629914, and the Global Collaboration Initiative (GCI) Program of the Notre Dame International Office, University of Notre Dame.

\bibliographystyle{splncs03}
\bibliography{Reference}
\end{document}